\documentclass{midl} 
\usepackage{mwe} 
\usepackage{adjustbox}
\usepackage{array} 
\usepackage{booktabs}
\usepackage{floatrow}
\usepackage{array}
\jmlrvolume{-- Under Review}
\jmlryear{2025}
\jmlrworkshop{Full Paper -- MIDL 2025 submission}
\editors{Under Review for MIDL 2025}

\title[ViT-2SPN]{ViT-2SPN: Vision Transformer-based Dual-Stream Self-Supervised Pretraining Networks for Retinal OCT Classification}

\midlauthor{\Name{Mohammadreza Saraei\nametag{$^{1}$}} \Email{mrsaraei@arizona.edu}\\
\addr $^{1}$ Department of Electrical and Computer Engineering, The University of Arizona, Tucson, AZ, USA \\
\Name{Igor Kozak\nametag{$^{2}$}} \Email{ikozak@arizona.edu}\\
\addr $^{2}$ Department of Ophthalmology and Vision Science, University of Arizona, Tucson, AZ, USA \AND
\Name{Eung-Joo Lee\midljointauthortext{Corresponding Author}\nametag{$^{3}$}} \Email{eungjoolee@arizona.edu}\\
\addr $^{3}$ Department of Electrical and Computer Engineering, The University of Arizona, Tucson, AZ, USA \\
}

\begin{document}

\maketitle

\begin{abstract}
Optical Coherence Tomography (OCT) is a non-invasive imaging modality essential for diagnosing various eye diseases. Despite its clinical significance, developing OCT-based diagnostic tools faces challenges, such as limited public datasets, sparse annotations, and privacy concerns. Although deep learning has made progress in automating OCT analysis, these challenges remain unresolved. To address these limitations, we introduce the Vision Transformer-based Dual-Stream Self-Supervised Pretraining Network (ViT-2SPN), a novel framework designed to enhance feature extraction and improve diagnostic accuracy. ViT-2SPN employs a three-stage workflow: Supervised Pretraining, Self-Supervised Pretraining (SSP), and Supervised Fine-Tuning. The pretraining phase leverages the OCTMNIST dataset (97,477 unlabeled images across four disease classes) with data augmentation to create dual-augmented views. A Vision Transformer (ViT-Base) backbone extracts features, while a negative cosine similarity loss aligns feature representations. Pretraining is conducted over 50 epochs with a learning rate of 0.0001 and momentum of 0.999. Fine-tuning is performed on a stratified 5.129\% subset of OCTMNIST using 10-fold cross-validation. ViT-2SPN achieves a mean AUC of 0.93, accuracy of 0.77, precision of 0.81, recall of 0.75, and an F1 score of 0.76, outperforming existing SSP-based methods. These results underscore the robustness and clinical potential of ViT-2SPN in retinal OCT classification. The code is available in \href{https://github.com/mrsaraei/ViT-2SPN.git}{https://github.com/mrsaraei/ViT-2SPN.git}.

\end{abstract}


\section{Introduction}
Medical imaging is a cornerstone of modern healthcare, enabling a detailed exploration of human anatomy and function. Imaging modalities, such as Optical Coherence Tomography (OCT), facilitate accurate diagnosis and monitoring of diseases \cite{bordbar2025use}. OCT, a non-invasive imaging tool, has transformed ophthalmology by providing high-resolution cross-sectional retinal images, critical for detecting retinal diseases \cite{subhedar2023review}. Using low-coherence light and interferometry, OCT generates axial (A-scan) and surface (B-scan) images \cite{patil2020development, patil2021algorithm}. However, the increasing volume of OCT scans challenges manual analysis, leading to the need for computer-assisted diagnostic (CAD) systems \cite{akpinar2024artificial} to improve efficiency and reduce errors \cite{albelwi2022survey}.

Deep Learning (DL), a subset of artificial intelligence, has revolutionized CAD systems by achieving human-level precision in disease classification \cite{chou2023current, heinke2024cross} and lesion detection under various conditions \cite{parmar2024artificial, saraei2023attention}. Techniques, such as Convolutional Neural Networks (CNNs) and Vision Transformers (ViTs) \cite{dosovitskiy2020image}, have made significant strides. However, these methods are based on Supervised Learning (SL), which requires large annotated data sets that are costly and raise privacy concerns \cite{taleb20203d}. Transfer Learning (TL) mitigates such challenges by adapting pretrained models for medical tasks \cite{oquab2014learning}, although its effectiveness is limited by the domain gaps between natural and medical images \cite{shurrab2022self}.

\begin{figure*}[ht]
\centering
\includegraphics[width=0.9\linewidth]{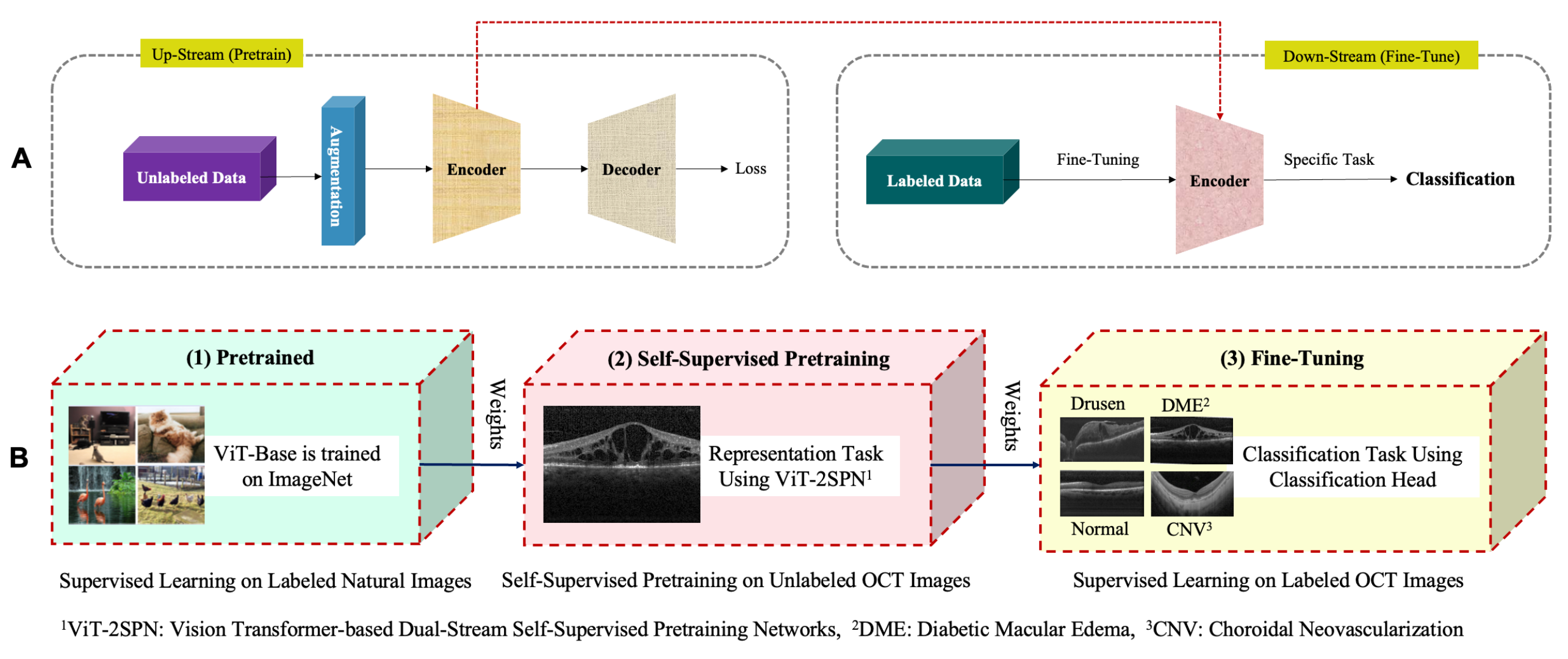}
\caption{(A) Self-supervised learning (SSL) pipeline for classification, (B) Self-supervised pertaining (SSP) approach for classification on the OCTMNIST dataset.}
\label{fig: Fig 1}
\end{figure*}

As depicted in Figure \ref{fig: Fig 1}A, Self-Supervised Learning (SSL) addresses these gaps by leveraging unlabeled data for pretraining, thereby reducing the reliance on manual annotations and improving scalability \cite{pani2024examining, zeng2024self, gui2024survey}. Fine-tuned SSL models have demonstrated exceptional classification and anomaly detection performance, effectively tackling data efficiency challenges in OCT analysis \cite{rani2024self}. Studies, such as those by Lee et al. and Bundele et al., further highlight the efficacy of combining pretraining, SSL, and fine-tuning for medical imaging tasks \cite{lee2022self, bundele2024evaluating}.

% Figure 1: The typical pipeline for applying self-supervised learning methods to downstream tasks (e.g., classification).

In this study, we present a method to improve the performance of retinal disease classification in OCT images by integrating pretraining, SSL, and supervised fine-tuning. Specifically, we propose a Dual-Stream Self-Supervised Pretraining (SSP) Network with a Vision Transformer (ViT-Base) backbone to improve feature extraction, detection accuracy, and robustness.

\section{Related Works}
Various DL models have been developed to improve the classification of retinal OCT, showing different levels of accuracy and efficiency. High-performing models such as Khalili et al.’s OCTNet (99.5\% accuracy) and Laouarem et al.’s hybrid CNN-ViT (99.77\% accuracy) exhibited impressive results. However, they were limited by high computational costs and overfitting on small datasets, reducing their practical use \cite{khalil2024octnet}, \cite{laouarem2024htc}. Prabha et al.’s lightweight model (97\%-98.08\% accuracy) balanced performance and efficiency but struggled with complex clinical conditions, highlighting a tradeoff between accuracy and scalability \cite{prabha2024rd}. Karthik et al.’s noise reduction technique improved accuracy by 2.44\%. However, its scope was narrow \cite{karthik2024deep}, while Dai et al.’s pretraining-based approach (95\% accuracy) showed strong domain-specific performance but lacked generalizability \cite{dai2024improving}.

On the other hand, SSL methods have garnered attention for improving classification performance, especially in data-scarce scenarios. Gholami et al. and Fang et al. demonstrated the effectiveness of SSL, with Gholami’s method for Macular Telangiectasia and Fang’s self-supervised patient-specific feature learning (SSPSF) achieving accuracies of 89.8\% and 97.74\%, respectively \cite{gholami2024self}, \cite{fang2022self}. SSL models also improved accuracy for complex conditions, such as Full-Thickness Macular Holes (FTMH) and Polypoidal Choroidal Vasculopathy (PCV) \cite{wheeler2024self}, \cite{wongchaisuwat2023automated}. Bundele et al. compared SSL models, finding that Momentum Contrast (MoCov3) achieved an AUC of 96.81\% and 79.96\% accuracy, while Barlow Twins and Nearest-Neighbor Contrastive Learning reached 97.11\% AUC, with accuracies of 77.28\% and 79.16\%, respectively \cite{bundele2024evaluating}. Despite their promise, SSL methods face challenges in designing robust pretraining strategies for clinical conditions and may perform sub-optimally on small or heterogeneous datasets.

\section{Methods}
As shown in Figure \ref{fig: Fig 1}B, our methodology consists of three stages to optimize the model to detect retinal diseases on OCT images. First, the model is initialized using ViT-base weights from ImageNet, transferring general visual knowledge to provide a strong foundation for feature extraction. Next, a dual-stream SSP architecture is applied to unlabeled OCT images, with a pretrained ViT-base backbone to capture rich feature extraction. Finally, the pretrained model is fine-tuned on a small set of labeled OCT images, with iterative hyperparameter optimization to improve classification accuracy and performance evaluation.

\subsection{Dataset}
\begin{figure*}[ht]
\centering
\includegraphics[width=0.9\linewidth]{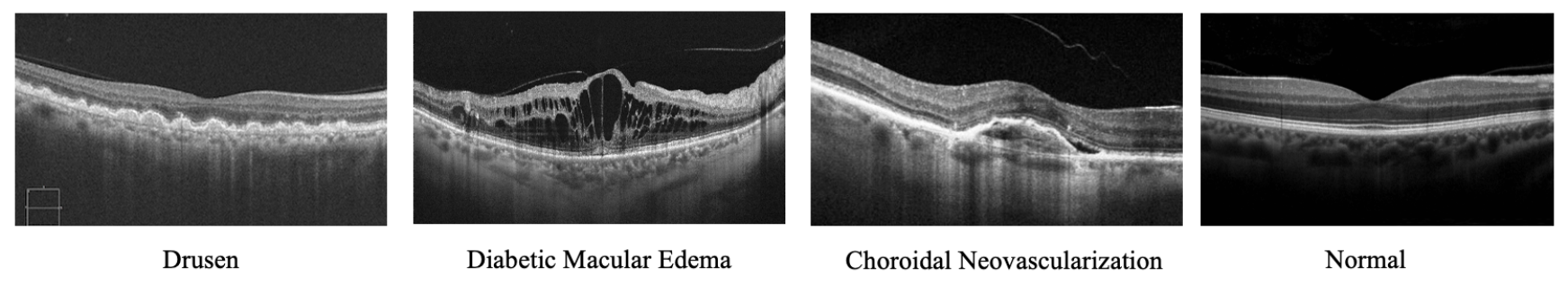}
\caption{Sample images of the OCTMNIST dataset for retinal disease classification.}
\label{fig: Fig 2}
\end{figure*}

We use the OCTMNIST dataset\footnote{\url{https://medmnist.com/}} to evaluate the efficacy of our training policy. OCTMNIST is derived from a publicly available retinal OCT dataset included in the MedMNISTv2 collection \cite{medmnistv2}. This dataset consists of four disease classes: Choroidal Neovascularization (CNV), Diabetic Macular Edema (DME), Drusen, and Normal (see Figure \ref{fig: Fig 2}). OCTMNIST is divided into 97,477, 10,832, and 21,816 images for training, validation, and test sets, respectively. Note that the reduction in image size does not significantly affect the accuracy of the classification \cite{lee2022self, yang2023medmnist}.

% Figure 2: The OCTMNIST Dataset

\subsection{Proposed Method}
\begin{figure*}[ht]
\centering
\includegraphics[width=1\linewidth]{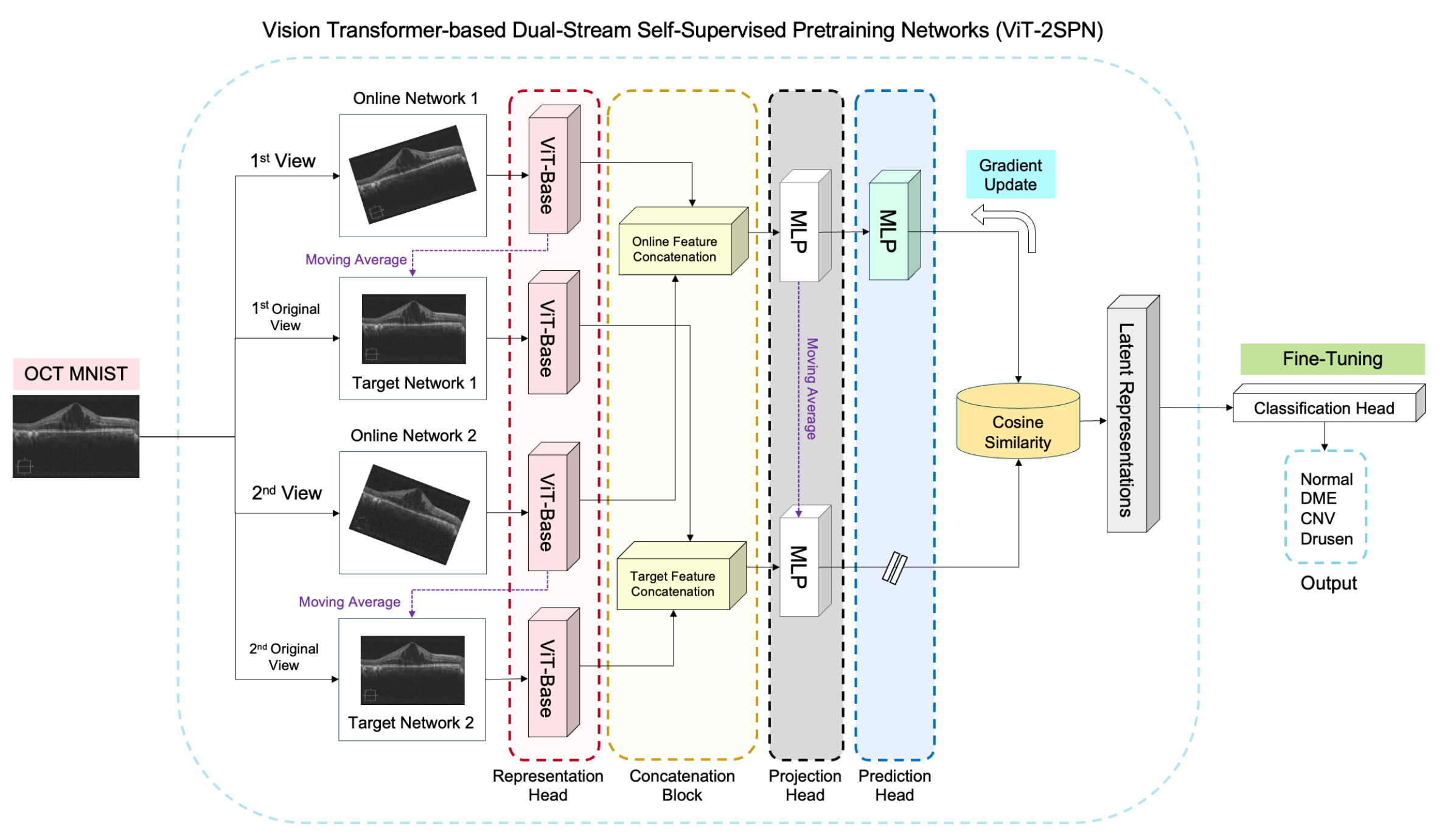}
\caption{Architecture of vision transformer-based dual-stream self-supervised pretraining networks (ViT-2SPN) for retinal OCT classification.}
\label{fig: Fig 3}
\end{figure*}

As demonstrated in Figure \ref{fig: Fig 3}, our proposed method, Vision Tranformer-based dual stream self-supervised pretraining networks (ViT-2SPN), is a novel framework designed for SSL and fine-tuning in OCT image classification tasks. The framework leverages dual-stream networks with a ViT-based backbone and contrastive learning objectives to generate robust latent feature representations. It uses pretraining on unlabeled OCTMNIST datasets, followed by fine-tuning for classification into clinical categories, such as Normal, DME, CNV, and Drusen.

% Figure 3: The ViT-2SPN architecture

The pretraining phase uses the OCTMNIST data set, which contains 97,477 unlabeled retinal OCT images. Data augmentation simulates variability and enhances generalization through random rotations, flips, grayscale transformations, color jitter, and normalization, creating dual augmented views as input for a dual-stream network. This architecture comprises online (encoder) and target (momentum encoder) streams, each using a ViT-base backbone pretrained on ImageNet to capture long dependencies vital for OCT imaging. The online stream processes one augmented view, while the target stream is updated using momentum-based moving averages for stability and refined representations.

Features from both streams are projected into a shared latent space through projection and prediction heads, aligning representations via a negative cosine similarity loss. This loss ensures the alignment of corresponding views while maintaining distinct representations for non-corresponding pairs. Specifically, the loss minimizes the cosine distance between online prediction features  $P_{online}$ and target projection features $Z_{target}$, as defined in Equation (\ref{eq:1}) (Appendix \ref{appendixA}).

\begin{equation}
\label{eq:1}
\
\mathcal{L}_{\text{total}} = \frac{-1}{N \cdot S} \sum_{s=1}^S \sum_{i=1}^N \frac{\mathbf{p}_{\text{online}}^{(i, s)} \cdot \mathbf{z}_{\text{target}}^{(i, s)}}{\|\mathbf{p}_{\text{online}}^{(i, s)}\| \|\mathbf{z}_{\text{target}}^{(i, s)}\|}
\
\end{equation}

This represents the total loss as the mean negative cosine similarity between the online network’s predicted features, \(\mathbf{p}_{\text{online}}^{(i, s)}\), and the target network’s projected representations, \(\mathbf{z}_{\text{target}}^{(i, s)}\), across all samples in a batch (\(N\)) and gradient accumulation steps (\(S\)).

The numerator, \(\mathbf{p}_{\text{online}}^{(i, s)} \cdot \mathbf{z}_{\text{target}}^{(i, s)}\), computes the dot product (similarity) between the two feature vectors. At the same time, the denominator normalizes this similarity by their magnitudes, ensuring that the comparison is scale-invariant. By averaging this loss across the entire data set processed in \(S\) steps, the method balances memory constraints and training stability while encouraging the online network to align with the representations of the target network. This is iteratively computed over epochs, and the momentum update ensures stable alignment between online and target features. Training employs a learning rate of \(10^{-4}\), a momentum of 0.999, and mixed precision techniques to optimize computational efficiency. Gradient accumulation over four steps further manages memory requirements for large batch sizes, and training is conducted for 50 epochs on a multi-GPU setup. The resulting pretrained weights encapsulating the learned representations are saved for subsequent fine-tuning. During fine-tuning, the pretrained backbone is adapted for supervised classification using a stratified subset of the OCTMNIST dataset. Data augmentation techniques, such as resizing, random rotations, flips, color jitter, and normalization are applied to improve generalization. Stratified 10-fold cross-validation ensures balanced class distribution across training and validation sets. A linear classification head, comprising a fully connected layer, dropout, batch normalization, and ReLU, maps latent features to diagnostic categories.

Training employs weighted cross-entropy loss to address class imbalances, optimized using the Adam optimizer (weight decay \(10^{-4}\)). A learning rate scheduler (ReduceLROnPlateau) adjusts the learning rate dynamically, while early stopping with three-epoch patience ensures efficient convergence. The best model, based on cross-validation ROC-AUC scores, is selected for evaluation. A separate test set of 500 samples ensures a robust and independent assessment.

The ViT-2SPN architecture integrates online and target networks, with momentum updates stabilizing feature alignment. ViT-base backbone highlighting critical features of the OCT image for pathology detection. Although reliance on labeled data may limit scalability, stratified cross-validation, and advanced optimization ensure a reliable performance evaluation.

\section{Experiments}

\subsection{Experimental Setup}
During the SSP phase, the model utilizes the unlabeled OCTMNIST dataset, which comprises 97,477 training samples. The training process is conducted with a mini-batch size of 128, a learning rate of 0.0001, and a momentum rate of 0.999, spanning a total of 50 epochs. The ViT-base architecture, pretrained on the ImageNet dataset, is employed as the backbone. In the fine-tuning phase, the model leverages 5.129\% of the labeled OCTMNIST dataset, following a 10-fold cross-validation strategy. Each fold consists of 4,500 training samples and 500 validation samples, with an additional 500 samples reserved for testing. The fine-tuning process is carried out using a batch size of 16, the same learning rate from the pretraining phase, a dropout rate of 0.5, and 50 epochs (see Appendix \ref{appendixB}).

\subsection{Results}
In Table \ref{tab:ssl_comparison}, we present an evaluation of the SSP models in the OCTMNIST dataset, showcasing the superior performance of ViT-2SPN across all key metrics. ViT-2SPN consistently outperforms prominent models, including Bootstrap Your Own Latent (BYOL) \cite{grill2020bootstrap}, Momentum Contrast (MoCo) \cite{he2020momentum} and its variants \cite{chen2020improved, chen2021empirical}, Simple Framework for Contrastive Learning (SimCLR) \cite{chen2020simple} and its updated version \cite{chen2020big}, Swapping Assignments Between Views (SwAV) \cite{caron2020unsupervised}, and Simple Siamese Representation Learning (SimSiam) \cite{chen2020simsiam}. The evaluation metrics include mean Area Under the Curve (mAUC), accuracy, precision, F1-score, and recall.

% Table 2: Performance comparison of self-supervised pretraining (SSP) models on the balanced OCTMNIST dataset.
\begin{table}[ht]
\centering
\caption{Performance comparison of SSP models on the OCTMNIST balanced dataset.}
\label{tab:ssl_comparison}
\begin{tabular}{lcccccc}
\toprule
\textbf{SSP Model} & \textbf{mAUC} & \textbf{Accuracy} & \textbf{Precision} & \textbf{F1-Score} & \textbf{Recall} \\ 
\midrule
BYOL & 0.660 & 0.54 & 0.58 & 0.53 & 0.54 \\ 
MoCo & 0.870 & 0.64 & 0.62 & 0.61 & 0.64 \\ 
SimCLR & 0.860 & 0.70 & 0.69 & 0.69 & 0.68 \\ 
SwAV & 0.863 & 0.66 & 0.67 & 0.66 & 0.66 \\ 
SimCLRv2 & 0.872 & 0.65 & 0.64 & 0.65 & 0.65 \\ 
MoCov2 & 0.870 & 0.68 & 0.68 & 0.68 & 0.69 \\ 
MoCov3 & 0.872 & 0.62 & 0.64 & 0.62 & 0.62 \\ 
SimSiam & 0.868 & 0.70 & 0.70 & 0.70 & 0.70 \\ 
ViT-2SPN (Ours) & \textbf{0.930} & \textbf{0.77} & \textbf{0.81} & \textbf{0.76} & \textbf{0.75} \\ 
\bottomrule
\end{tabular}
\end{table}

\begin{figure*}[ht]
\centering
\includegraphics[width=0.95\linewidth]{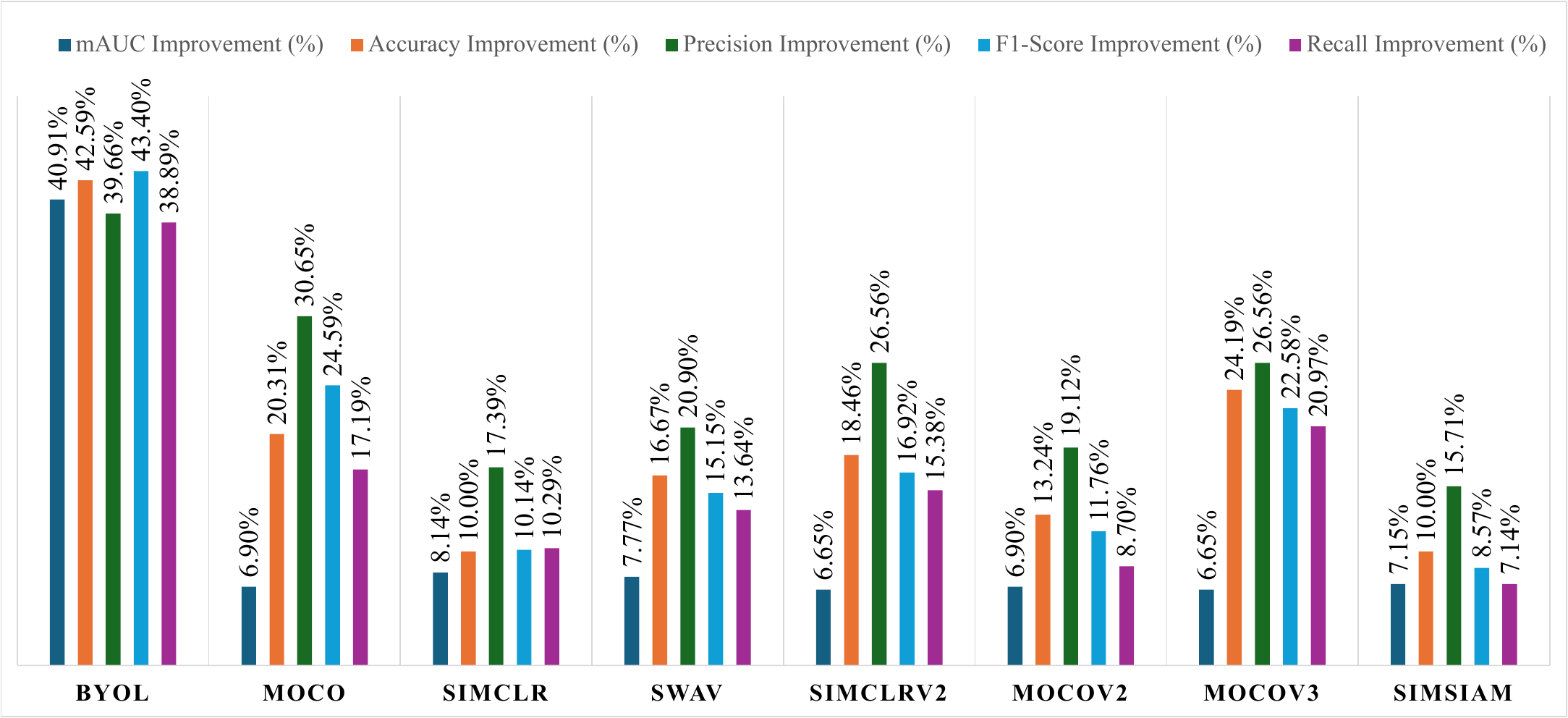}
\caption{Performance improvements for ViT-2SPN on the balanced OCTMNIST dataset.}
\label{fig: Fig 4}
\end{figure*}

\begin{figure*}[ht]
\centering
\includegraphics[width=0.95\linewidth]{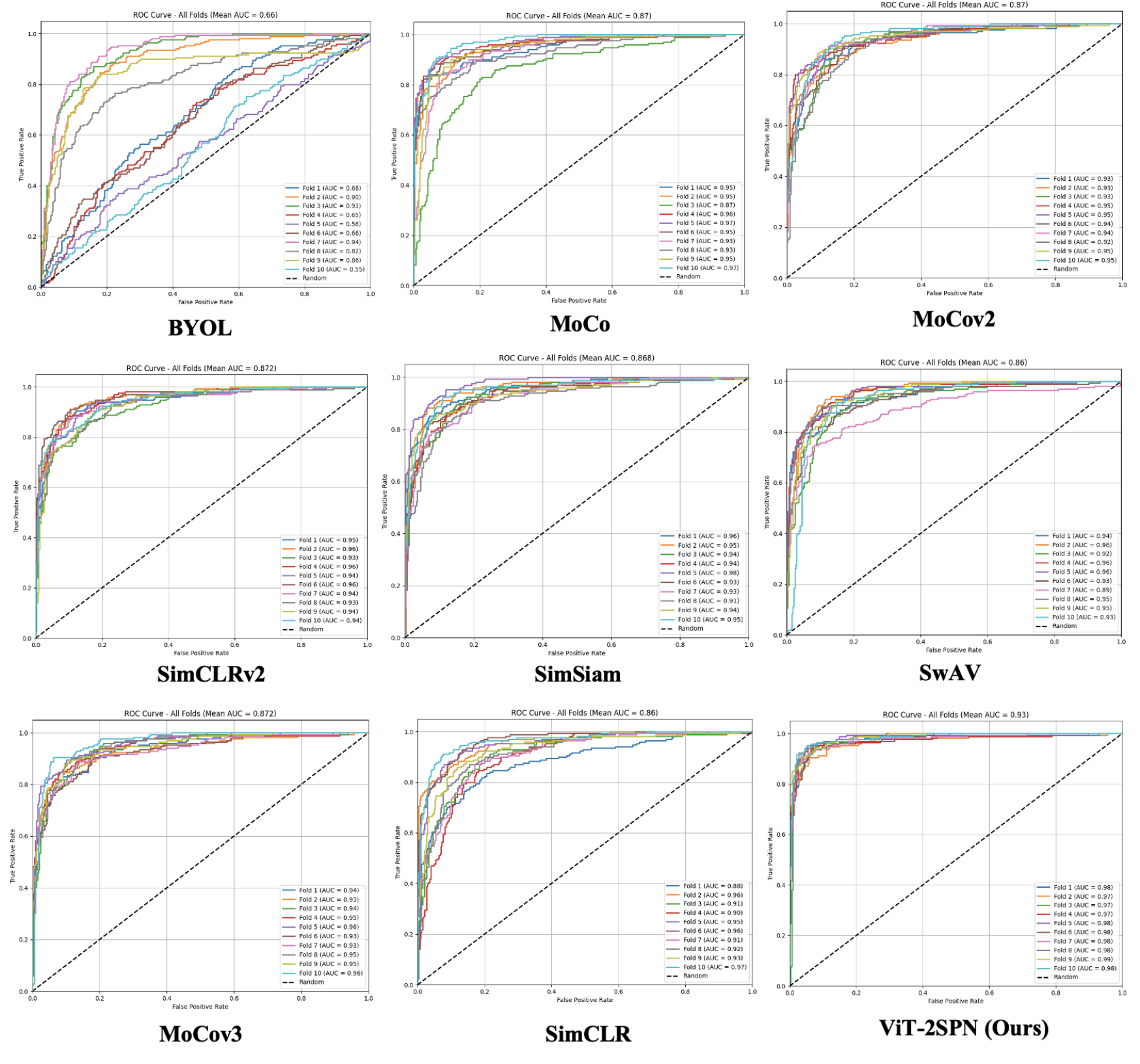}
\caption{Comparison of AUC curves for various SSP models on the balanced OCTMNIST dataset.}
\label{fig: Fig 5}
\end{figure*}

As illustrated in Figure \ref{fig: Fig 4}, ViT-2SPN demonstrates superior capability in modeling complex patterns in OCT imaging, leading to improved evaluation metrics for classifying retinal diseases. For instance, as shown in Figure \ref{fig: Fig 5}, ViT-2SPN achieves a mAUC of 0.930, significantly outperforming SimCLRv2 and MoCov3, both of which score 0.872, as well as BYOL, which has a score of 0.660. Furthermore, ViT-2SPN achieves a classification accuracy of 0.77, exceeding SimCLR and SimSiam, both at 0.70 and BYOL, which scores 0.54. This model also demonstrates a precision of 0.81, indicating a lower rate of false positives compared to SimCLR and SimSiam (both at 0.70) and BYOL (0.58). Its F1-score of 0.76 and recall of 0.75 further underscore its strong sensitivity and balance, outperforming SimCLR and SimSiam (both at 0.70) and BYOL, which lag with scores in the range of 0.53–0.54. These findings affirm the superior generalization capabilities of ViT-2SPN relative to competing models (see the confusion matrix evaluation in Appendix \ref{appendixB}).

While models such as MoCo variants deliver consistent but moderate performance, SwAV and SimCLRv2 show promise in mAUC yet underperform in other metrics. Despite their competitive accuracy and precision, SimCLR and SimSiam remain limited in their ability to match ViT-2SPN’s advanced capabilities, revealing the constraints of traditional contrastive learning methods.

% Figure 4: Achieved performance improvements for ViT-2SPN on the OCTMNIST dataset.

% Figure 5: Comparison of AUC curves and confusion matrices for self-supervised pretraining models on the balanced OCTMNIST dataset.

\section{Conclusion}
This study presents ViT-2SPN, a Vision Transformer-based dual stream self-supervised pretraining network, designed to address challenges such as limited annotated datasets and concerns for patient privacy. Using SSL and an innovative pretraining strategy on the OCTMNIST dataset, ViT-2SPN enhances the extraction of features and achieves significant improvements in the classification of retinal diseases. The performance of the model, with a mean AUC of 0.93 and an accuracy of 0.77, precision of 0.81, recall of 0.75, and an F1-score of 0.76, demonstrates its robustness and effectiveness compared to other SSP-based approaches. The integration of ViT-base backbone, dual-stream data augmentation, and a carefully designed training pipeline ensure that ViT-2SPN not only captures complex features but also generalizes well to unseen data. These results underscore the robustness and clinical potential of ViT-2SPN in OCT image classification. ViT-2SPN represents a promising step forward in leveraging DL for non-invasive and accurate ophthalmic disease diagnosis, contributing to improved patient outcomes and advancing the field of OCT imaging. However, challenges remain, including computational cost, inference time, and the need to optimize computational efficiency. Future work will focus on integrating knowledge distillation techniques to enhance scalability, extending the framework to larger and more diverse datasets, exploring its applicability to other imaging modalities, and addressing current limitations to maximize its clinical utility.

% \clearpage  % Acknowledgements, references, and appendix do not count toward the page limit (if any)
% Acknowledgments---Will not appear in the anonymized version
% \midlacknowledgments{We thank a bunch of people.}

%\section{Code Availability}

%The code is available at \url{https://github.com/mrsaraei/ViT-2SPN.git}.

\bibliography{midl-samplebibliography}

\begin{thebibliography}{39}
\providecommand{\natexlab}[1]{#1}
\providecommand{\url}[1]{\texttt{#1}}
\expandafter\ifx\csname urlstyle\endcsname\relax
  \providecommand{\doi}[1]{doi: #1}\else
  \providecommand{\doi}{doi: \begingroup \urlstyle{rm}\Url}\fi

\bibitem[Akpinar et~al.(2024)Akpinar, Sengur, Faust, Tong, Molinari, and Acharya]{akpinar2024artificial}
Muhammed~Halil Akpinar, Abdulkadir Sengur, Oliver Faust, Louis Tong, Filippo Molinari, and U~Rajendra Acharya.
\newblock Artificial intelligence in retinal screening using oct images: A review of the last decade (2013-2023).
\newblock \emph{Computer Methods and Programs in Biomedicine}, page 108253, 2024.

\bibitem[Albelwi(2022)]{albelwi2022survey}
Saleh Albelwi.
\newblock Survey on self-supervised learning: auxiliary pretext tasks and contrastive learning methods in imaging.
\newblock \emph{Entropy}, 24\penalty0 (4):\penalty0 551, 2022.

\bibitem[Bordbar et~al.(2025)Bordbar, Bhatnagar, and Weng]{bordbar2025use}
Darius~D Bordbar, Anshul Bhatnagar, and Christina~Y Weng.
\newblock Use of home optical coherence tomography for retinal diseases.
\newblock \emph{International Ophthalmology Clinics}, 65\penalty0 (1):\penalty0 41--46, 2025.

\bibitem[Bundele et~al.(2024)Bundele, {\c{C}}al, Kargi, Sar{\i}ta{\c{s}}, Tez{\"o}ren, Ghaderi, and Lensch]{bundele2024evaluating}
Valay Bundele, O{\u{g}}uz~Ata {\c{C}}al, Bora Kargi, Karahan Sar{\i}ta{\c{s}}, K{\i}van{\c{c}} Tez{\"o}ren, Zohreh Ghaderi, and Hendrik Lensch.
\newblock Evaluating self-supervised learning in medical imaging: A benchmark for robustness, generalizability, and multi-domain impact.
\newblock \emph{arXiv preprint arXiv:2412.19124}, 2024.

\bibitem[Caron et~al.(2020)Caron, Misra, Mairal, Goyal, Bojanowski, and Joulin]{caron2020unsupervised}
Mathilde Caron, Ishan Misra, Julien Mairal, Priya Goyal, Piotr Bojanowski, and Armand Joulin.
\newblock Unsupervised learning of visual features by contrasting cluster assignments.
\newblock \emph{Advances in neural information processing systems}, 33:\penalty0 9912--9924, 2020.

\bibitem[Chen et~al.(2020{\natexlab{a}})Chen, Kornblith, Norouzi, and Hinton]{chen2020simple}
Ting Chen, Simon Kornblith, Mohammad Norouzi, and Geoffrey Hinton.
\newblock A simple framework for contrastive learning of visual representations.
\newblock In \emph{International conference on machine learning}, pages 1597--1607. PMLR, 2020{\natexlab{a}}.

\bibitem[Chen et~al.(2020{\natexlab{b}})Chen, Kornblith, Swersky, Norouzi, and Hinton]{chen2020big}
Ting Chen, Simon Kornblith, Kevin Swersky, Mohammad Norouzi, and Geoffrey~E Hinton.
\newblock Big self-supervised models are strong semi-supervised learners.
\newblock \emph{Advances in neural information processing systems}, 33:\penalty0 22243--22255, 2020{\natexlab{b}}.

\bibitem[Chen and He(2020)]{chen2020simsiam}
Xinlei Chen and Kaiming He.
\newblock Exploring simple siamese representation learning.
\newblock \emph{arXiv preprint arXiv:2011.10566}, 2020.

\bibitem[Chen et~al.(2020{\natexlab{c}})Chen, Fan, Girshick, and He]{chen2020improved}
Xinlei Chen, Haoqi Fan, Ross Girshick, and Kaiming He.
\newblock Improved baselines with momentum contrastive learning.
\newblock \emph{arXiv preprint arXiv:2003.04297}, 2020{\natexlab{c}}.

\bibitem[Chen et~al.(2021)Chen, Xie, and He]{chen2021empirical}
Xinlei Chen, Saining Xie, and Kaiming He.
\newblock An empirical study of training self-supervised vision transformers.
\newblock In \emph{Proceedings of the IEEE/CVF international conference on computer vision}, pages 9640--9649, 2021.

\bibitem[Chou et~al.(2023)Chou, Kale, Lanzetta, Aslam, Barratt, Danese, Eldem, Eter, Gale, Korobelnik, et~al.]{chou2023current}
Yu-Bai Chou, Aditya~U Kale, Paolo Lanzetta, Tariq Aslam, Jane Barratt, Carla Danese, Bora Eldem, Nicole Eter, Richard Gale, Jean-Fran{\c{c}}ois Korobelnik, et~al.
\newblock Current status and practical considerations of artificial intelligence use in screening and diagnosing retinal diseases: Vision academy retinal expert consensus.
\newblock \emph{Current opinion in ophthalmology}, 34\penalty0 (5):\penalty0 403--413, 2023.

\bibitem[Dai et~al.(2024)Dai, Yang, Yue, and Chen]{dai2024improving}
Hao Dai, Yaliang Yang, Xian Yue, and Shen Chen.
\newblock Improving retinal oct image classification accuracy using medical pre-training and sample replication methods.
\newblock \emph{Biomedical Signal Processing and Control}, 91:\penalty0 106019, 2024.

\bibitem[Dosovitskiy(2020)]{dosovitskiy2020image}
Alexey Dosovitskiy.
\newblock An image is worth 16x16 words: Transformers for image recognition at scale, 2020.

\bibitem[Fang et~al.(2022)Fang, Guo, He, and Li]{fang2022self}
Leyuan Fang, Jiahuan Guo, Xingxin He, and Muxing Li.
\newblock Self-supervised patient-specific features learning for oct image classification.
\newblock \emph{Medical \& Biological Engineering \& Computing}, 60\penalty0 (10):\penalty0 2851--2863, 2022.

\bibitem[Gholami et~al.(2024)Gholami, Scheppke, Kshirsagar, Wu, Dodhia, Bonelli, Leung, Sallo, Muldrew, Jamison, et~al.]{gholami2024self}
Shahrzad Gholami, Lea Scheppke, Meghana Kshirsagar, Yue Wu, Rahul Dodhia, Roberto Bonelli, Irene Leung, Ferenc~B Sallo, Alyson Muldrew, Catherine Jamison, et~al.
\newblock Self-supervised learning for improved optical coherence tomography detection of macular telangiectasia type 2.
\newblock \emph{JAMA ophthalmology}, 142\penalty0 (3):\penalty0 226--233, 2024.

\bibitem[Grill et~al.(2020)Grill, Strub, Altch{\'e}, Tallec, Richemond, Buchatskaya, Doersch, Avila~Pires, Guo, Gheshlaghi~Azar, et~al.]{grill2020bootstrap}
Jean-Bastien Grill, Florian Strub, Florent Altch{\'e}, Corentin Tallec, Pierre Richemond, Elena Buchatskaya, Carl Doersch, Bernardo Avila~Pires, Zhaohan Guo, Mohammad Gheshlaghi~Azar, et~al.
\newblock Bootstrap your own latent-a new approach to self-supervised learning.
\newblock \emph{Advances in neural information processing systems}, 33:\penalty0 21271--21284, 2020.

\bibitem[Gui et~al.(2024)Gui, Chen, Zhang, Cao, Sun, Luo, and Tao]{gui2024survey}
Jie Gui, Tuo Chen, Jing Zhang, Qiong Cao, Zhenan Sun, Hao Luo, and Dacheng Tao.
\newblock A survey on self-supervised learning: Algorithms, applications, and future trends.
\newblock \emph{IEEE Transactions on Pattern Analysis and Machine Intelligence}, 2024.

\bibitem[He et~al.(2020)He, Fan, Wu, Xie, and Girshick]{he2020momentum}
Kaiming He, Haoqi Fan, Yuxin Wu, Saining Xie, and Ross Girshick.
\newblock Momentum contrast for unsupervised visual representation learning.
\newblock In \emph{Proceedings of the IEEE/CVF conference on computer vision and pattern recognition}, pages 9729--9738, 2020.

\bibitem[Heinke et~al.(2024)Heinke, Zhang, Broniarek, Michalska-Ma{\l}ecka, Elsner, Galang, Deussen, Warter, Kalaw, Nagel, et~al.]{heinke2024cross}
Anna Heinke, Haochen Zhang, Krzysztof Broniarek, Katarzyna Michalska-Ma{\l}ecka, Wyatt Elsner, Carlo Miguel~B Galang, Daniel~N Deussen, Alexandra Warter, Fritz Kalaw, Ines Nagel, et~al.
\newblock Cross-instrument optical coherence tomography-angiography (octa)-based prediction of age-related macular degeneration (amd) disease activity using artificial intelligence.
\newblock \emph{Scientific Reports}, 14\penalty0 (1):\penalty0 27085, 2024.

\bibitem[Karthik and Mahadevappa(2024)]{karthik2024deep}
Karri Karthik and Manjunatha Mahadevappa.
\newblock Deep learning with adaptive convolutions for classification of retinal diseases via optical coherence tomography.
\newblock \emph{Image and Vision Computing}, 146:\penalty0 105044, 2024.

\bibitem[Khalil et~al.(2024)Khalil, Mehmood, Kim, and Kim]{khalil2024octnet}
Irshad Khalil, Asif Mehmood, Hyunchul Kim, and Jungsuk Kim.
\newblock Octnet: A modified multi-scale attention feature fusion network with inceptionv3 for retinal oct image classification.
\newblock \emph{Mathematics}, 12\penalty0 (19):\penalty0 3003, 2024.

\bibitem[Laouarem et~al.(2024)Laouarem, Kara-Mohamed, Bourennane, and Hamdi-Cherif]{laouarem2024htc}
Ayoub Laouarem, Chafia Kara-Mohamed, El-Bay Bourennane, and Aboubekeur Hamdi-Cherif.
\newblock Htc-retina: a hybrid retinal diseases classification model using transformer-convolutional neural network from optical coherence tomography images.
\newblock \emph{Computers in Biology and Medicine}, 178:\penalty0 108726, 2024.

\bibitem[Lee and Lee(2022)]{lee2022self}
Joohyung Lee and Eung-Joo Lee.
\newblock Self-supervised pre-training improves fundus image classification for diabetic retinopathy.
\newblock In \emph{Real-time image processing and deep learning 2022}, volume 12102, pages 193--198. SPIE, 2022.

\bibitem[Oquab et~al.(2014)Oquab, Bottou, Laptev, and Sivic]{oquab2014learning}
Maxime Oquab, Leon Bottou, Ivan Laptev, and Josef Sivic.
\newblock Learning and transferring mid-level image representations using convolutional neural networks.
\newblock In \emph{Proceedings of the IEEE conference on computer vision and pattern recognition}, pages 1717--1724, 2014.

\bibitem[Pani and Chawla(2024)]{pani2024examining}
Kaliprasad Pani and Indu Chawla.
\newblock Examining the quality of learned representations in self-supervised medical image analysis: a comprehensive review and empirical study.
\newblock \emph{Multimedia Tools and Applications}, pages 1--31, 2024.

\bibitem[Parmar et~al.(2024)Parmar, Surico, Singh, Romano, Salati, Spadea, Musa, Gagliano, Mori, and Zeppieri]{parmar2024artificial}
Uday Pratap~Singh Parmar, Pier~Luigi Surico, Rohan~Bir Singh, Francesco Romano, Carlo Salati, Leopoldo Spadea, Mutali Musa, Caterina Gagliano, Tommaso Mori, and Marco Zeppieri.
\newblock Artificial intelligence (ai) for early diagnosis of retinal diseases.
\newblock \emph{Medicina}, 60\penalty0 (4):\penalty0 527, 2024.

\bibitem[Patil et~al.(2020)Patil, Mahajan, Balamurugan, Arulmozhivarman, and Makkar]{patil2020development}
Kranti Patil, Anurag Mahajan, S~Balamurugan, P~Arulmozhivarman, and Roshan Makkar.
\newblock Development of signal processing algorithm for optical coherence tomography.
\newblock In \emph{2020 International Conference on Communication and Signal Processing (ICCSP)}, pages 1283--1287. IEEE, 2020.

\bibitem[Patil et~al.(2021)Patil, Mahajan, Subramani, Pachiyappan, and Makkar]{patil2021algorithm}
Kranti Patil, Anurag Mahajan, Balamurugan Subramani, Arulmozhivarman Pachiyappan, and Roshan Makkar.
\newblock Algorithm for b-scan image reconstruction in optical coherence tomography.
\newblock \emph{Pertanika J. Sci. Technol.}, 29\penalty0 (1):\penalty0 533--545, 2021.

\bibitem[Prabha et~al.(2024)Prabha, Venkatesan, Fathimal, Nithiyanantham, and Kirubha]{prabha2024rd}
A~Jeya Prabha, C~Venkatesan, M~Sameera Fathimal, KK~Nithiyanantham, and SP~Angeline Kirubha.
\newblock Rd-oct net: hybrid learning system for automated diagnosis of macular diseases from oct retinal images.
\newblock \emph{Biomedical Physics \& Engineering Express}, 10\penalty0 (2):\penalty0 025033, 2024.

\bibitem[Rani et~al.(2024)Rani, Kumar, Gupta, Sachdeva, Mittal, and Kumar]{rani2024self}
Veenu Rani, Munish Kumar, Aastha Gupta, Monika Sachdeva, Ajay Mittal, and Krishan Kumar.
\newblock Self-supervised learning for medical image analysis: a comprehensive review.
\newblock \emph{Evolving Systems}, pages 1--27, 2024.

\bibitem[Saraei and Liu(2023)]{saraei2023attention}
Mohammadreza Saraei and Sidong Liu.
\newblock Attention-based deep learning approaches in brain tumor image analysis: A mini review.
\newblock \emph{Frontiers in Health Informatics}, 12:\penalty0 1--9, 2023.

\bibitem[Shurrab and Duwairi(2022)]{shurrab2022self}
Saeed Shurrab and Rehab Duwairi.
\newblock Self-supervised learning methods and applications in medical imaging analysis: A survey.
\newblock \emph{PeerJ Computer Science}, 8:\penalty0 e1045, 2022.

\bibitem[Subhedar and Mahajan(2023)]{subhedar2023review}
Jahida Subhedar and Anurag Mahajan.
\newblock A review on recent work on oct image classification for disease detection.
\newblock In \emph{2022 OPJU International Technology Conference on Emerging Technologies for Sustainable Development (OTCON)}, pages 1--6. IEEE, 2023.

\bibitem[Taleb et~al.(2020)Taleb, Loetzsch, Danz, Severin, Gaertner, Bergner, and Lippert]{taleb20203d}
Aiham Taleb, Winfried Loetzsch, Noel Danz, Julius Severin, Thomas Gaertner, Benjamin Bergner, and Christoph Lippert.
\newblock 3d self-supervised methods for medical imaging.
\newblock \emph{Advances in neural information processing systems}, 33:\penalty0 18158--18172, 2020.

\bibitem[Wheeler et~al.(2024)Wheeler, Hunter, Garcia, Li, Thomson, Hunter, and Mehanian]{wheeler2024self}
Timothy~William Wheeler, Kaitlyn Hunter, Patricia~Anne Garcia, Henry Li, Andrew~Clark Thomson, Allan Hunter, and Courosh Mehanian.
\newblock Self-supervised contrastive learning improves machine learning discrimination of full thickness macular holes from epiretinal membranes in retinal oct scans.
\newblock \emph{PLOS Digital Health}, 3\penalty0 (8):\penalty0 e0000411, 2024.

\bibitem[Wongchaisuwat et~al.(2023)Wongchaisuwat, Thamphithak, Watunyuta, and Wongchaisuwat]{wongchaisuwat2023automated}
Nida Wongchaisuwat, Ranida Thamphithak, Poom Watunyuta, and Papis Wongchaisuwat.
\newblock Automated classification of polypoidal choroidal vasculopathy and wet age-related macular degeneration by spectral domain optical coherence tomography using self-supervised learning.
\newblock \emph{Procedia Computer Science}, 220:\penalty0 1003--1008, 2023.

\bibitem[Yang et~al.(2023{\natexlab{a}})Yang, Shi, Wei, Liu, Zhao, Ke, Pfister, and Ni]{medmnistv2}
Jiancheng Yang, Rui Shi, Donglai Wei, Zequan Liu, Lin Zhao, Bilian Ke, Hanspeter Pfister, and Bingbing Ni.
\newblock Medmnist v2-a large-scale lightweight benchmark for 2d and 3d biomedical image classification.
\newblock \emph{Scientific Data}, 10\penalty0 (1):\penalty0 41, 2023{\natexlab{a}}.

\bibitem[Yang et~al.(2023{\natexlab{b}})Yang, Shi, Wei, Liu, Zhao, Ke, Pfister, and Ni]{yang2023medmnist}
Jiancheng Yang, Rui Shi, Donglai Wei, Zequan Liu, Lin Zhao, Bilian Ke, Hanspeter Pfister, and Bingbing Ni.
\newblock Medmnist v2-a large-scale lightweight benchmark for 2d and 3d biomedical image classification.
\newblock \emph{Scientific Data}, 10\penalty0 (1):\penalty0 41, 2023{\natexlab{b}}.

\bibitem[Zeng et~al.(2024)Zeng, Abdullah, and Sumari]{zeng2024self}
Xiangrui Zeng, Nibras Abdullah, and Putra Sumari.
\newblock Self-supervised learning framework application for medical image analysis: a review and summary.
\newblock \emph{BioMedical Engineering OnLine}, 23\penalty0 (1):\penalty0 107, 2024.

\end{thebibliography}

\newpage
\appendix

\section{Proof of Theorem \ref{eq:1}} \label{appendixA}
The loss function used in our training is based on cosine similarity, which ensures that the online prediction features $P_{online}$ align closely with the corresponding target projection features $Z_{target}$. 

Let:

\begin{itemize}
    \item \(\mathbf{z}_{\text{online}}^{(i)} \in \mathbb{R}^d\) be the online network’s projected representation for view \(i\).
    \item \(\mathbf{p}_{\text{online}}^{(i)} \in \mathbb{R}^d\) be the online network’s predicted feature for view \(i\).
    \item \(\mathbf{z}_{\text{target}}^{(i)} \in \mathbb{R}^d\) be the target network’s projected representation for view \(i\).
\end{itemize}

For two augmented views of the same image \(x\):

\begin{enumerate}
    \item Online representations:
    \begin{equation}
    \label{eq:2}
    \
    \mathbf{z}_{\text{online}} = f_{\text{online}}(x_1) \quad \text{and} \quad \mathbf{p}_{\text{online}} = g_{\text{online}}(\mathbf{z}_{\text{online}})
    \
    \end{equation}
    \item Target representations:
    \begin{equation}
    \label{eq:3}
    \
    \mathbf{z}_{\text{target}} = f_{\text{target}}(x_2)
    \
    \end{equation}
    \noindent where \(f(\cdot)\) is the encoder (backbone network), and \(g(\cdot)\) is the projection head.
\end{enumerate}

The loss for a single pair of representations is as follows:

\begin{equation}
\label{eq:4}
\
\mathcal{L}_{\text{pair}} = - \text{CosineSimilarity}(\mathbf{p}_{\text{online}}, \mathbf{z}_{\text{target}}) = - \frac{\mathbf{p}_{\text{online}} \cdot \mathbf{z}_{\text{target}}}{\|\mathbf{p}_{\text{online}}\| \|\mathbf{z}_{\text{target}}\|}
\
\end{equation}

Since we compute the mean over the batch, the total loss for a batch of size \(N\) is:

\begin{equation}
\label{eq:5}
\
\mathcal{L}_{\text{batch}} = - \frac{1}{N} \sum_{i=1}^N \frac{\mathbf{p}_{\text{online}}^{(i)} \cdot \mathbf{z}_{\text{target}}^{(i)}}{\|\mathbf{p}_{\text{online}}^{(i)}\| \|\mathbf{z}_{\text{target}}^{(i)}\|}
\
\end{equation}

In our implementation, gradient accumulation is used to handle memory constraints by splitting the computation across \(S\) steps:

\begin{equation}
\label{eq:6}
\
\mathcal{L}_{\text{accumulated}} = \frac{1}{S} \sum_{s=1}^S \mathcal{L}_{\text{batch}}^{(s)}
\
\end{equation}

The target network is updated using an exponential moving average of the weights of the online network:

\begin{equation}
\label{eq:7}
\
\theta_{\text{target}} \gets m \cdot \theta_{\text{target}} + (1 - m) \cdot \theta_{\text{online}}
\
\end{equation}

\noindent where \(m\), momentum, is a hyperparameter.
\\

The overall loss can be expressed as follows:

\begin{equation}
\label{eq:8}
\
\mathcal{L}_{\text{total}} = \frac{-1}{N \cdot S} \sum_{s=1}^S \sum_{i=1}^N \frac{\mathbf{p}_{\text{online}}^{(i, s)} \cdot \mathbf{z}_{\text{target}}^{(i, s)}}{\|\mathbf{p}_{\text{online}}^{(i, s)}\| \|\mathbf{z}_{\text{target}}^{(i, s)}\|}
\
\end{equation}

\section{Supplementary Materials \label{appendixB}}

% Table 1: The experimental setup of our proposed model (ViT-2SPN) for the OCTMNIST dataset
\begin{table}[ht]
    \centering
    \caption{Experimental setup and implementation of ViT-2SPN on the balanced OCTMNIST dataset.}
    \label{tab:experimental_setup}
    \begin{tabular}{lcc|lcc}
        \toprule
        \textbf{Self-Supervised Pretraining} & \textbf{Value} & & \textbf{Fine-Tuning} & \textbf{Value} \\ 
        \midrule
        Dataset & OCTMNIST & & Dataset & OCTMNIST \\ 
        Train & 97,477 & & 10-Fold Train & 4,500 \\ 
        Mini-Batch Size & 128 & & 10-Fold Validation & 500 \\ 
        Learning Rate & 0.0001 & & Test & 500 \\ 
        Momentum Rate & 0.999 & & Batch Size & 16 \\ 
        Epoch & 50 & & Epoch & 50 \\
        Backbone & ViT-base & & Dropout Rate & 0.5 \\
        \bottomrule
    \end{tabular}
\end{table}

% Figure 6: Comparison of confusion matrices for self-supervised pretraining models on the balanced OCTMNIST dataset.
\begin{figure*}[ht]
\centering
\includegraphics[width=1\linewidth]{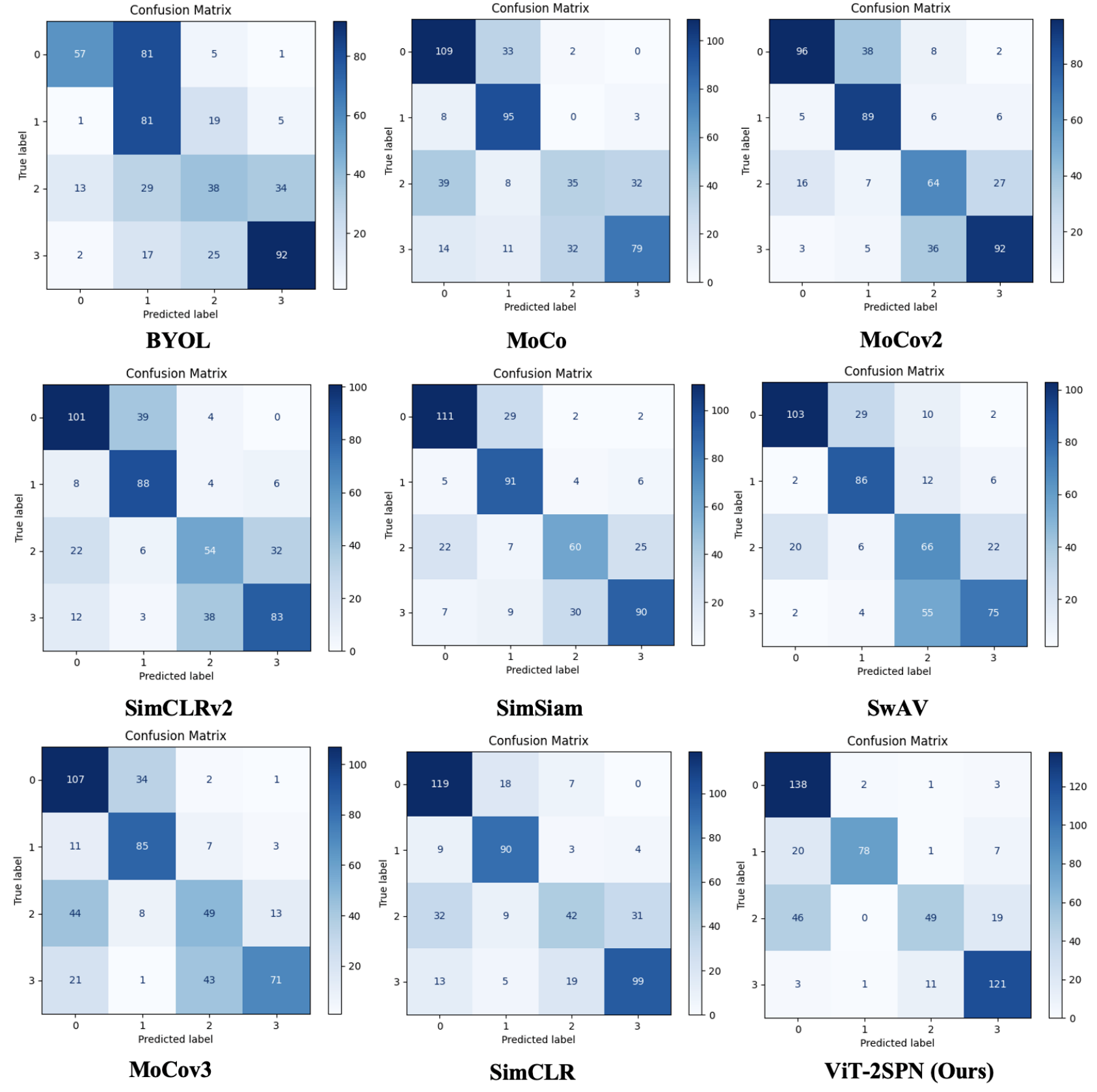}
\caption{Comparison of confusion matrices for different SSP models on the balanced OCTMNIST dataset.}
\label{fig: Fig 6}
\end{figure*}

\end{document}